\documentclass[journal]{IEEEtran}
\usepackage{algorithm}
\usepackage{algorithmic}
\usepackage{multirow}
\usepackage{bbm}
\usepackage{amsmath,amssymb,mathrsfs}
\usepackage{graphicx} 
\usepackage{float} 
\usepackage{subfigure} 
\usepackage{amsthm}
\usepackage[switch]{lineno} 
\usepackage{amsmath,amssymb}
\usepackage{algorithm}
\usepackage{algorithmic}
\usepackage{graphicx}
\usepackage{comment}
\usepackage{color}
\usepackage{subfigure}
\usepackage{multirow}
\usepackage{algorithm}
\usepackage{algorithmic}
\usepackage{wrapfig}
\usepackage{authblk}
\usepackage{booktabs}
\usepackage{hyperref}
\usepackage{xcolor}
\hypersetup{
	colorlinks=true,
	urlcolor=cyan,
}

\ifCLASSINFOpdf
\else
\fi
\begin{document}
	\newtheorem{theorem}{Theorem}
	\newtheorem{lemma}{Lemma}
	\newtheorem{property}{Property}	
	\newtheorem{theorem2}{Theorem}
	\newtheorem{lemma2}{Lemma}
	\newtheorem{property2}{Property}	
	\title{Point Cloud Registration-Driven Robust Feature Matching for 3D Siamese Object Tracking}
	\newcommand*\samethanks[1][\value{footnote}]{\footnotemark[#1]}
	\author[ ]{
		Haobo Jiang, Kaihao Lan, Le Hui, Guangyu Li, Jin Xie, and Jian Yang
		\thanks{Haobo Jiang, Kaihao Lan, Le Hui, Guangyu Li, Jin Xie, and Jian Yang are with PCA Lab, Key Lab of Intelligent Perception and Systems for High-Dimensional Information of Ministry of Education, and Jiangsu Key Lab of Image and Video Understanding for Social Security, School of Computer Science and Engineering, Nanjing University of Science and Technology, Nanjing 210094, China (e-mail: jiang.hao.bo@njust.edu.cn; lkh@njust.edu.cn; le.hui@njust.edu.cn; guangyu.li2017@njust.edu.cn; csjxie@njust.edu.cn; csjyang@njust.edu.cn).
		}}
	
	\markboth{Journal of \LaTeX\ Class Files,~Vol.~14, No.~8, August~2015}%
	{Shell \MakeLowercase{\mathrm{et al.}}: Bare Demo of IEEEtran.cls for IEEE Journals}
	\maketitle
	
\begin{abstract}
Learning robust feature matching between the template and search area is crucial for 3D Siamese tracking. 
The core of Siamese feature matching is how to assign high feature similarity on the corresponding points between the template and search area for precise object localization. 
In this paper, we propose a novel point cloud registration-driven Siamese tracking framework, with the intuition that spatially aligned corresponding points (via 3D registration) tend to achieve consistent feature representations. 
Specifically, our method consists of two modules, including a tracking-specific nonlocal registration module and a registration-aided Sinkhorn template-feature aggregation module. 
The registration module targets at the precise spatial alignment between the template and search area. The tracking-specific spatial distance constraint is proposed to refine the cross-attention weights in the nonlocal module for discriminative feature learning. 
Then, we use the weighted SVD to compute the rigid transformation between the template and search area, and align them to achieve the desired spatially aligned corresponding points. 
For the feature aggregation model, we formulate the feature matching between the transformed template and search area as an optimal transport problem and utilize the Sinkhorn optimization to search for the outlier-robust matching solution. 
Also, a registration-aided spatial distance map is built to improve the matching robustness in indistinguishable regions (e.g., smooth surface). 
Finally, guided by the obtained feature matching map, we aggregate the target information from the template into the search area to construct the target-specific feature, which is then fed into a CenterPoint-like detection head for object localization. 
Extensive experiments on KITTI, NuScenes and Waymo datasets verify the effectiveness of our proposed method. 
	\end{abstract}
	
	\section{Introduction}
Visual object tracking serves as the key component in a variety of computer vision applications, such as autonomous driving~\cite{luo2018fast}, robot vision~\cite{comport2004robust}, and augmented reality~\cite{yan2020pointasnl}. 
With the development of inexpensive LiDAR sensors, more research has turned to 3D object tracking at the point cloud level and achieved substantial progress.
In general, given the annotated 3D bounding box of the target in the first frame, the tracker aims to predict the positions of the target in the remaining scanning frames.
Compared to trackers using 2D images, point cloud-based trackers are inherently more robust in challenging situations, e.g., illumination and weather changes. 
However, reliable 3D object tracking in the real world remains a challenging problem in the presence of the sparse scene, object occlusion, and LiDAR noise.

\begin{figure}[t]
	\centering
	\includegraphics[width=1.0\columnwidth]{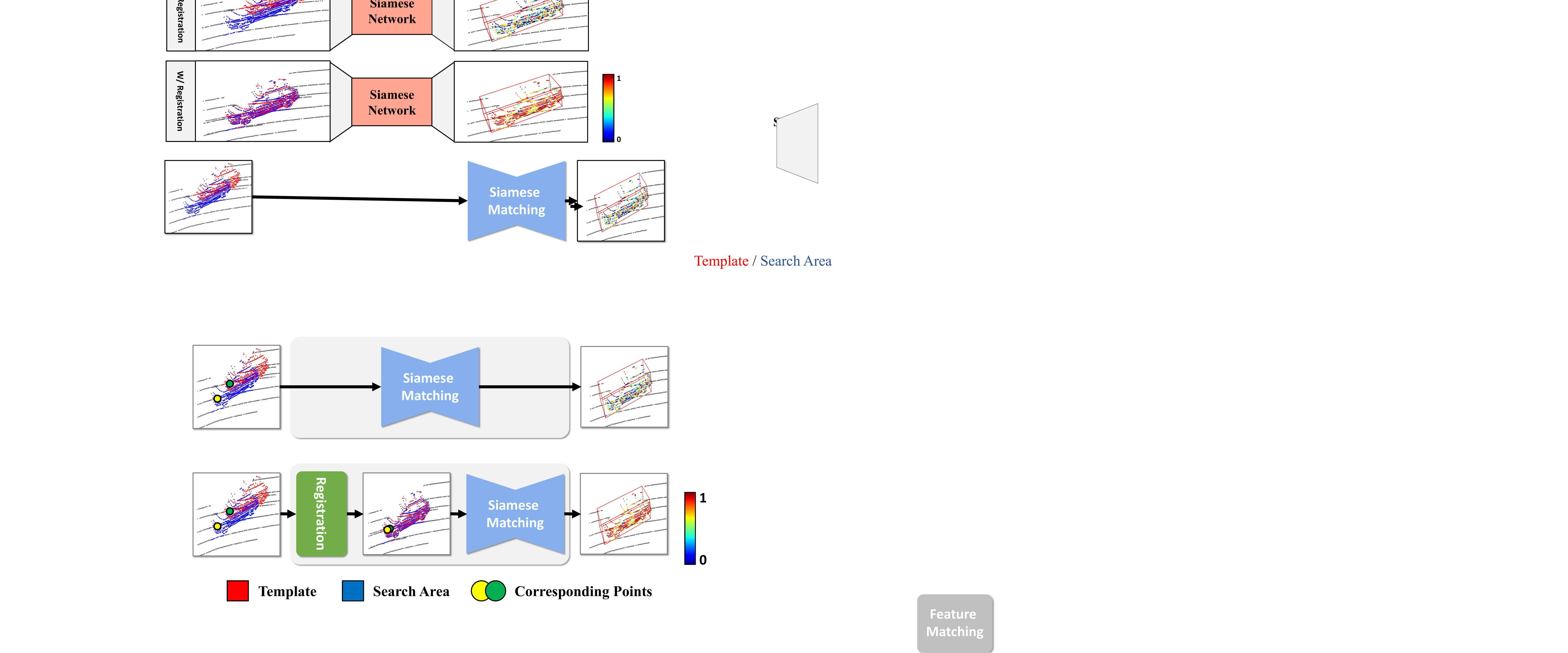}
	\caption{Feature matchings between the original template and the search area (top row) and between  the transformed template and the search area (bottom row). 
		Compared to the former, by spatially aligning the corresponding points  (green and yellow points) via registration, their feature representation tends to be consistent, resulting in much higher feature similarity for object localization.}
	\label{fig:motivation}
\end{figure}

In recent years, 3D Siamese object tracking has attracted increasing research interest due to the success of 2D Siamese tracking and the development of deep learning in 3D vision field.
The core of the Siamese tracker is to obtain reliable feature matching between the target template and the search area, thereby robustly distinguishing the target from the search area. 
As a pioneer, SC3D~\cite{giancola2019leveraging} uses the Kalman filtering to generate a set of target candidates and utilizes a shape completion-enhanced Siamese network to measure the feature similarity between the candidates and templates.
The target candidate with the largest similarity score is then selected as the target location. 
To improve tracking efficiency and enable end-to-end training, P2B~\cite{qi2020p2b} jointly performs the target proposal generation and verification for object localization. 
Particularly, guided by the reliable feature matching between the template and search area, the target information in the template is aggregated into the search area to generate the target-specific feature for position and rotation angle regression. 
Following this pipeline, BAT~\cite{zheng2021box} further boosts the feature matching by constructing more discriminative size-aware and part-aware BoxCloud features. 
Furthermore, to handle 3D tracking in sparse scenes, V2B~\cite{hui20213d} utilizes the shape-aware feature learning for voxel-to-BEV target detection. 
In summary, the tracking performance of the Siamese trackers introduced above relies heavily on the reliable feature matching, regardless of SC3D using the candidate with the largest similarity score as the target location, or P2B, BAT, and V2B exploiting the feature similarity to guide the target-specific feature learning. 

In this paper, we propose a simple yet effective 3D Siamese tracking framework that focuses on leveraging the point cloud registration for robust feature matching.
The motivation is that the feature extraction backbone in Siamese network is essentially a projection function from the 3D coordinate space to the feature space.
Thus, the spatially aligned corresponding points (with nearly consistent coordinates) tend to achieve consistent feature representations (as shown in Fig.\ref{fig:motivation}). 

Specifically, our framework consists of two modules, including a Tracking-Specific Nonlocal Registration module (TSNR) and a Registration-aided Sinkhorn Feature Aggregation module (RSFA). 
Taking as input the template and search area, the registration module aims to accurately predict the rigid transformation for their spatial alignment by alternately performing the nonlocal feature extraction, the inlier classification, and the weighted SVD. 
To achieve a discriminative feature representation, TSNR exploits the tracking-specific spatial distance constraint between the corresponding points to refine the cross-attention weights in the nonlocal module. 
The intuition of such spatial constraint is that the distance between the corresponding points from two consecutive frames is usually limited due to the small scanning interval of LiDAR. 
After transforming the template with the estimated rigid transformation, a PointNet++ backbone~\cite{qi2017pointnet++} is employed to extract the feature embeddings of the transformed template and the search area.
Different from previous Siamese trackers that directly uses the Cosine distance to measure the feature similarity, RSFA formulates the feature matching (between the transformed template and the search area) as an optimal transport problem and utilizes the Sinkhorn optimization to search for the outlier-robust matching solution. 
Furthermore, to improve the robustness of the feature matching in indistinguishable regions (e.g., car-door surfaces), a registration-aided spatial distance map is also built to regularize the feature matching similarity. 
Consequently, we exploit the refined feature matching map to guide the target information (defined in the template) aggregation into the search area and form the target-specific search-area feature for object localization via a CenterPoint-like detection head~\cite{yin2021center,hui20213d}. 
Extensive experiments on KITTI, NuScenes and Waymo benchmark datasets verify the effectiveness of our proposed method. To summarize, our main contributions are listed as follows:
\begin{itemize}
	\item We propose a novel 3D registration-driven Siamese tracking framework, where the robust feature matching can be obtained by spatially aligning the corresponding points between the template and search area.
	\item We design an effective tracking-specific nonlocal registration module for reliable registration between the template and the search area during 3D tracking.
	\item We propose a novel registration-aided feature aggregation module, where the registration-based spatial distance map is proposed for feature matching refinement. 
	\item Compared to current state-of-the-art methods, our proposed tracker can obtain outstanding performance on extensive benchmark datasets. 
\end{itemize}
	
\section{Related Work}
\subsection{3D single object tracking} 
Due to the inherent robustness of point clouds to real-world tracking challenges such as illumination changes and less-textured scenes, more research has moved from RGB-based 2D tracking to 3D tracking that primarily utilizes LiDAR-scanned point clouds for object tracking. 
SC3D~\cite{giancola2019leveraging} pioneeringly leverages shape completion to enhance the feature representation of the template and the target candidates (obtained by Kalman filtering), and selects the target candidate with the largest feature similarity to the template as the target localization. 
P2B~\cite{qi2020p2b} constructs the discriminative target-specific feature for object identification by integrating the template information into the search area.  
Then, the VoteNet~\cite{qi2019deep} is employed to determine the object position in the search area.  
3DSiamRPN~\cite{fang20203d} proposes to exploit the region proposal network to obtain the proposal and scores for object localization. 
BAT~\cite{zheng2021box} proposes to construct the bounding box-aware feature descriptor for a more robust feature matching. 
In addition, to handle the tracking task in the sparse point cloud scenes, V2B~\cite{hui20213d}  utilizes the discriminative shape-aware template embedding for feature aggregation and replaces the VoteNet with a voxel-to-BEV detector for target location. 
Recently, inspired by the success of transformers, PTT~\cite{shan2021ptt} employs the transformer-based self-attention mechanism for feature learning.  
PTTR~\cite{zhou2021pttr} fuses the self-attention and cross-attention mechanisms in transformer for robust target-specific feature learning and achieves impressive tracking accuracy. 

\subsection{3D multi-object tracking} 
Most multi-object trackers such as \cite{wu20213d,shenoi2020jrmot,kim2021eagermot} first exploit a 3D detector \cite{shi2020points,shi2019pointrcnn} to determine the targets to be tracked in all frames (i.e., tracking-by-detection). 
Then, the data association among the detected objects in the consecutive frames is used for object trajectories prediction. 
\cite{weng2019baseline} proposes to utilize the 3D Kalman filter for tracking state estimation and then exploit the Hungarian algorithm for matching the detected objects. 
\cite{wang2020joint} further uses a GNN module for the object relationship modeling and focuses on jointly optimizing the object detection and the data association. 
CenterPoint \cite{yin2021center} finds the centers of objects and regresses other tracking properties using the keypoint detector and further refines these estimated object locations via additional point features based on the predicted  position. 
Other methods~\cite{wu20213d,patil2019h3d} also show enlightening results. 

\begin{figure*}[t]
	\centering
	\includegraphics[width=1.0\textwidth]{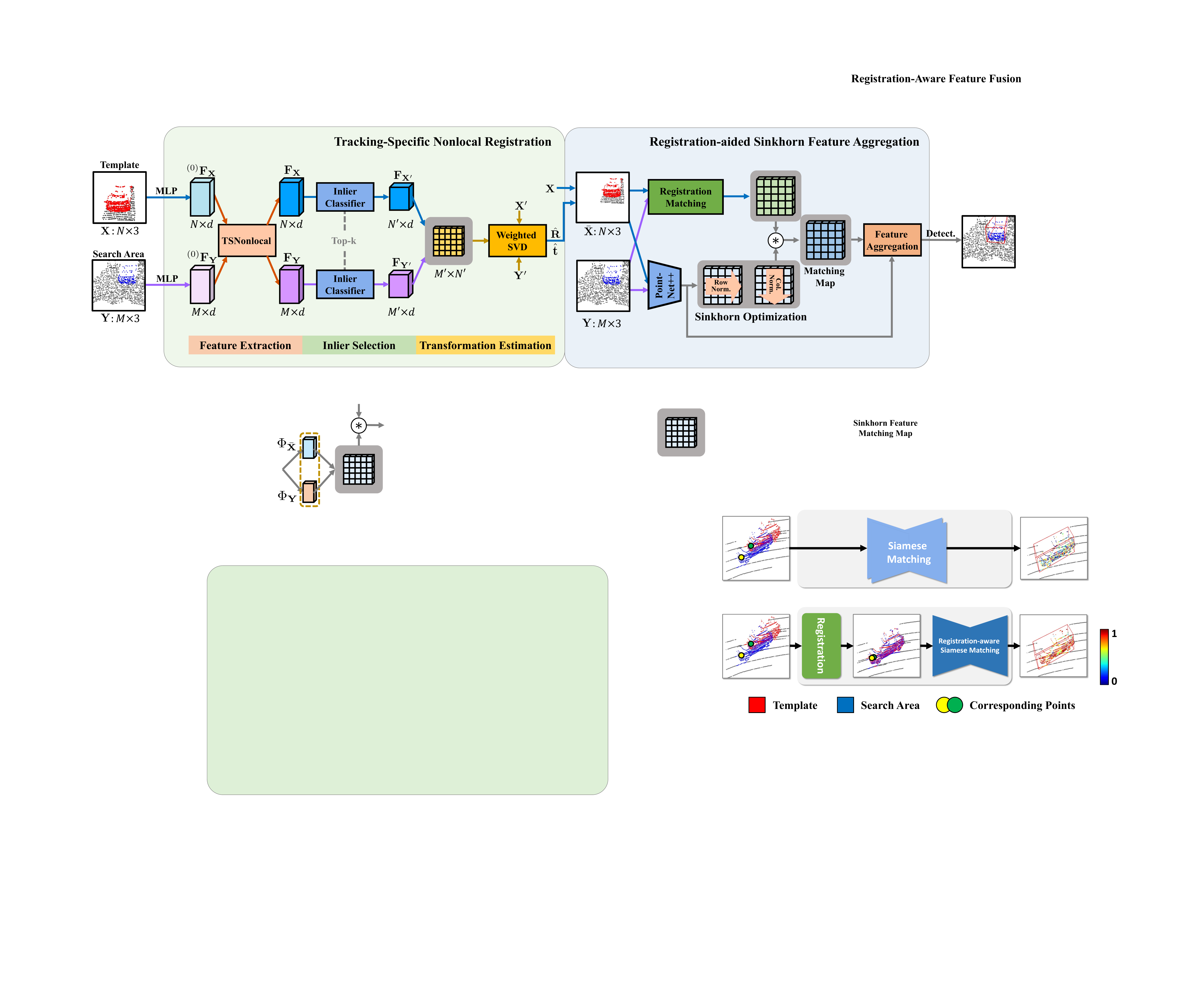}
	\caption{The pipeline of our point cloud registration-driven Siamese tracking framework, which contains two modules: 
		(a) Tracking-specific nonlocal registration: Given the template $\mathbf{X}$ and search area $\mathbf{Y}$, the proposed tracking-specific nonlocal (TSNonlocal) module first extracts the geometric feature for each point. An inlier classifier is then built to select points with high inlier probabilities for transformation estimation $\{\hat{\mathbf{R}},\hat{\mathbf{t}}\}$ via the weighted SVD.  
		(b) Registration-aided Sinkhorn feature aggregation: We first extract the PointNet++ features on the transformed template $\bar{\mathbf{X}}=\{ {\hat{\mathbf{R}}}{\mathbf{x}}_i+\hat{\mathbf{t}}\}$ and the search area $\mathbf{Y}$. 
		Then, we combine the Sinkhorn optimization and the registration-aided spatial distance map for robust feature matching map construction.
		Finally, we exploit the obtained matching map for target-specific feature aggregation, where the aggregated feature is then used for object localization through a CenterPoint-like detection head.}
	\label{fig:pipeline}
\end{figure*}

\subsection{Point cloud registration} 
Taking as input a pair of source and target point clouds, the point cloud registration aims to recover their potential rigid transformation for spatial alignment. 
For traditional registration methods, iterative closest point algorithm (ICP)~\cite{besl1992method} iteratively performs correspondence estimation and least-squares optimization for transformation estimation. 
Other ICP variants, such as Go-ICP~\cite{yang2013go} and Trimmed ICP~\cite{chetverikov2002trimmed}, improve ICP in terms of the local optima problem and the partially-overlapping registration. 
For learning-based methods, DCP~\cite{wang2019deep} utilizes feature similarity maps to generate pseudo-correspondences and then computes the rigid transformation using singular value decomposition (SVD). 
To handle partially overlapped point clouds, PRNet~\cite{wang2019prnet} improves the DCP by iteratively detecting the key points and identifying the correspondences in a self-supervised manner. 
Furthermore, by fusing the geometric characteristic and the deep feature, IDAM~\cite{li2020iterative} constructs the distance-aware soft correspondence to achieve reliable correspondence estimation. 
PointNetLK~\cite{aoki2019pointnetlk} exploits the differentiable Lucas-Kanade algorithm to iteratively optimize the feature distance for rigid transformation calculation. 
By formulating the 3D registration task as the probability matching problem, DeepGMR~\cite{yuan2020deepgmr} realizes the point cloud alignment by minimizing their KL divergence. 
In addition, inspired by the success of reinforcement learning, CEMNet~\cite{jiang2021sampling}, PlanPR~\cite{jiang2021planning} and RegAgent~\cite{bauer2021reagent} formulate the point cloud registration problem as the Markov decision process and utilizes the model-based planning decision and model-free proximal policy optimization algorithm, respectively, to search for the optimal rigid transformation. 

\section{Approach} 
\subsection{Problem Setting}
In point cloud-based single object tracking, given the 3D bounding box (BBox) $\mathbf{b}_1$ of the target in the first frame, the tracker aims to predict the BBoxes $\{\hat{\mathbf{b}}_{t>1}\}$ of the target in remaining scanning frames. 
The BBox can be parameterized with 7 elements, including the object center coordinate $(x,y,z)$, the object size $(l, w, h)$ and the heading angle $\theta$ (we assume the rotation of the object is just around the \textit{z}-axis). 
Since the size of BBox keeps consistent with $\mathbf{b}_1$ during tracking, we needn't infer the object size $(l, w, h)$. 
Following the Siamese network paradigm, we aim to localize the target (defined by the template) in the search area frame by frame. 
Specifically, in frame $t$, we first construct the template $\mathbf{X}=\left\{\mathbf{x}_i \in \mathbb{R}^3\mid i=1,...,N\right\}$ and search area $\mathbf{Y}=\left\{\mathbf{y}_j \in \mathbb{R}^3 \mid j=1,...,M\right\}$ by cropping frame $t-1$ with $\hat{\mathbf{b}}_{t-1}$ and  frame $t$ with enlarged $\hat{\mathbf{b}}_{t-1}$, respectively. 
Then, taking as input the template and search area, the Siamese tracker predicts the BBox $\hat{\mathbf{b}}_t$ including the center coordinate $(x,y,z)$ and the rotation angle $\theta$ for object localization. 

Fig.~\ref{fig:pipeline} illustrates our overall tracking framework containing two modules: a tracking-specific nonlocal registration module (Sec.~\ref{sec:tanr}) and a registration-aided Sinkhorn feature aggregation module (Sec.~\ref{fusion}). 
The former focuses on spatially aligning the corresponding points between the template and search area, and the latter performs robust feature matching between the two for object identification.

\subsection{Tracking-Specific Nonlocal Registration} 
\label{sec:tanr}
The tracking-specific nonlocal registration module formulates the template and search area $\left\{\mathbf{X},\mathbf{Y}\right\}$ as the partially overlapped source and target point cloud pair in 3D registration paradigm, and aims to align $\mathbf{X}$ to $\mathbf{Y}$ by inferring their rigid transformation consisting of a rotation matrix $\mathbf{R} \in SO\left(3\right)$ and a translation vector $\mathbf{t}\in \mathbb{R}^3$. 

\begin{figure*}[t]
	\centering
	\includegraphics[width=2.0\columnwidth]{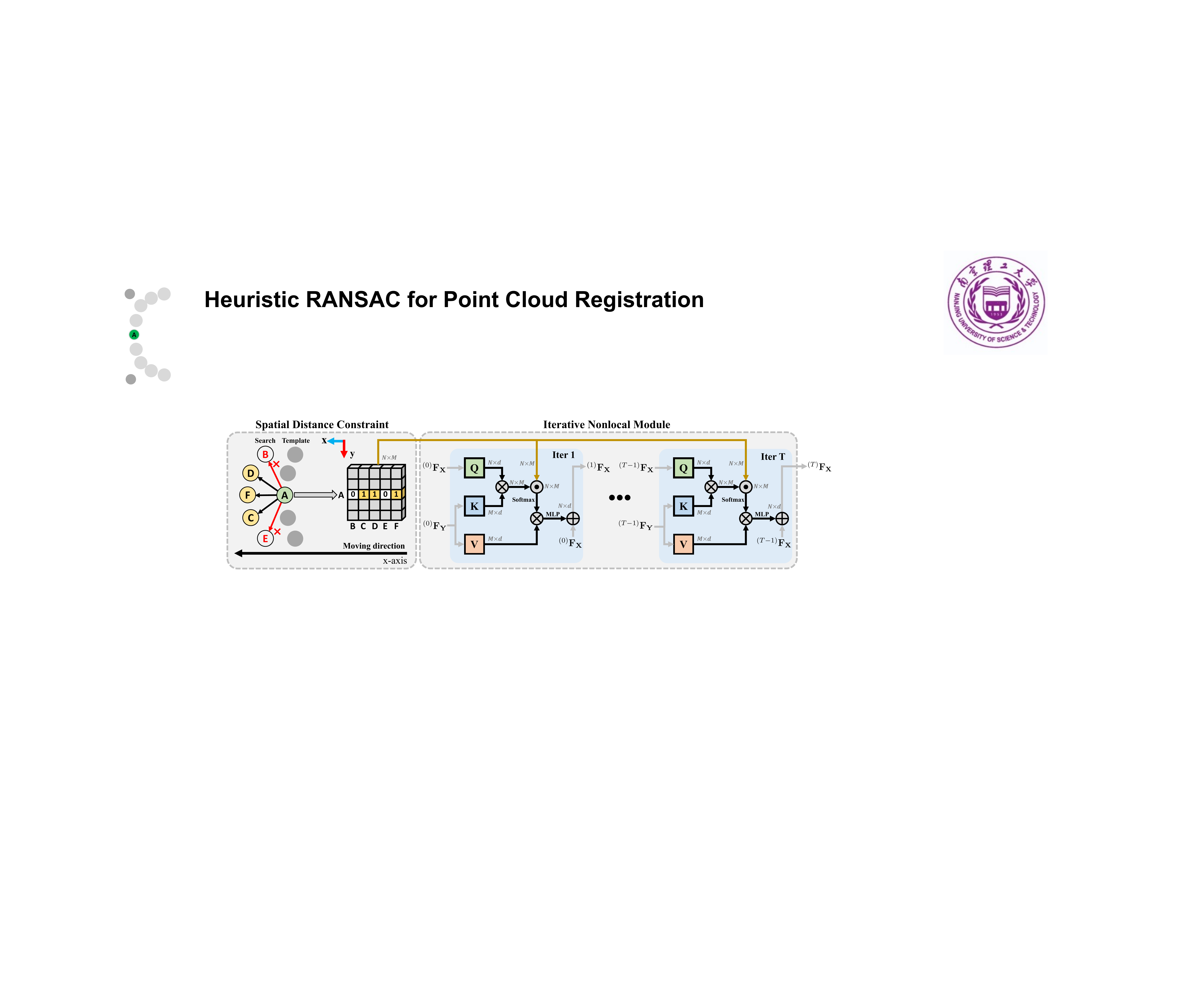}
	\caption{Tracking-specific iterative nonlocal (TSNonlocal) module. The left block illustrates the spatial distance constraint between the template and search area (top view without \textit{z}-axis). 
		The right block presents the feature enhancement process based on the iterative nonlocal module, where the spatial distance constraint is integrated into the $K$-$Q$ correlation map for regularizing the cross-attention weights. }
	\label{fig:tanonlocal}
\end{figure*}

\noindent\textbf{Tracking-specific nonlocal feature learning.}  
The feature learning network used in our registration module is the Tracking-Specific iterative Nonlocal module (TSNonlocal), which receives the template and search area as input and produces a geometric feature for each point. 
To realize robust registration performance,  our TSNonlocal module explicitly utilizes the spatial distance constaint between the corresponding points of the template and search area to learn a discriminative feature representation. 
As illustrated in Fig.~\ref{fig:tanonlocal}, TSNonlocal contains $T$ iterations for feature updating and the initial feature embeddings $\{{}^{(0)}\mathbf{F}_{\mathbf{x}_i}\in\mathbb{R}^d\}$ and $\{{}^{(0)}\mathbf{F}_{\mathbf{y}_j}\in\mathbb{R}^d\}$  are obtained by mapping each point $\mathbf{x}_i$ and $\mathbf{y}_j$ into a common feature space with a multi-layer perception (MLP).  
Then, the initial features are iteratively enhanced through the nonlocal message passing from the search area to the template ( $\mathbf{Y}\rightarrow \mathbf{X}$) and the template to the search area ($\mathbf{X}\rightarrow \mathbf{Y}$). 
To avoid repeating, we just describe the message passing about $\mathbf{Y}\rightarrow \mathbf{X}$. 
Specifically, in $t$-th iteration ($1 \leq t\leq T$), the enhanced point feature ${}^{(t)}\mathbf{F}_{{\mathbf{x}_i}}\in\mathbb{R}^d$ is obtained by the way that the query ${}^{(t)}\mathbf{Q}_{\mathbf{x}_i}={}^{(t)}\mathbf{W}_q{ }^{(t-1)} \mathbf{F}_{\mathbf{x}_{i}} \in\mathbb{R}^d$ retrieves the value ${}^{(t)}\mathbf{V}_{\mathbf{y}_j}={}^{(t)}\mathbf{W}_v{ }^{(t-1)} \mathbf{F}_{\mathbf{y}_{j}}\in\mathbb{R}^d$ with key ${}^{(t)}\mathbf{K}_{\mathbf{y}_j}={}^{(t)}\mathbf{W}_k{ }^{(t-1)} \mathbf{F}_{\mathbf{y}_{j}}\in\mathbb{R}^d$ (${}^{(t)}\mathbf{W}_q$, ${}^{(t)}\mathbf{W}_k$ and ${}^{(t)}\mathbf{W}_v$ are the learnable parameters): 
\begin{equation}\label{nonlocal_eq}
	{ }^{(t)} \mathbf{F}_{\mathbf{x}_{i}}={ }^{(t-1)} \mathbf{F}_{\mathbf{x}_{i}}+\operatorname{MLP}\Big(\sum_{j=1}^{M}{ }^{(t)} \alpha_{i, j} \beta_{i, j} {}^{(t)}\mathbf{V}_{\mathbf{y}_j}\Big),
\end{equation}
where the cross-attention weight ${ }^{(t)}\alpha_{i, j}$ is defined as the dot-product similarity between the query and key: 
\begin{equation}
	{ }^{(t)}\alpha_{i, j}=\operatorname{softmax}({ }^{(t)}\mathbf{Q}_{\mathbf{x}_i}^{\top(t)} \mathbf{K}_{\mathbf{y}_j} / \sqrt{d}). 
\end{equation}
Furthermore, we exploit a spatial distance constraint $\beta_{i, j}$ between the template and search area to regularize the cross-attention weight. 
The intuition of this spatial constraint lies on the observation that the distance between the corresponding points of the template and search area is limited due to the small LiDAR scanning interval and the relatively stable object motion. 
As depicted in Fig.~\ref{fig:tanonlocal}, the corresponding point of template point $\mathbf{A}$ tends to lie in its spatially close  points $\{\mathbf{D},\mathbf{F},\mathbf{C}\}$ in the search area while the far points $\{\mathbf{B},\mathbf{E}\}$ may be rejected.  
This spatial constraint-enhanced cross attention explicitly mitigates the unexpected message passing from the non-corresponding points $\{\mathbf{B},\mathbf{E}\}$ to $\mathbf{A}$, thereby assisting  more robust message passing and feature learning. 
In our implementation, we heuristically design the spatial distance constraint as below:
\begin{equation}\label{beta}
	\beta_{i j}=\mathbb{I}\left\{d_{ij}\leq1\right\},  d_{i j}=\frac{\left(\mathbf{y}_{j,y}-\mathbf{x}_{i,y}\right)^2}{a^2}+\frac{\left(\mathbf{y}_{j,z}-\mathbf{x}_{i,z}\right)^2}{b^2}, 
\end{equation}
where $\mathbf{x}_{i,y}$ and $\mathbf{x}_{i,z}$ denotes the \textit{y}-axis value (left/right direction) and \textit{z}-axis value (up/down direction)  of template point $\mathbf{x}_i$, respectively. 
For each template-point $\mathbf{x}_i$, its possible corresponding points in search area are defined as the points whose Euclidean distances to $\mathbf{x}_i$ fall in an ellipse area ($a$ and $b$ are the semi-major and semi-minor axes) in the \textit{yz}-plane (x-axis denotes the forward/backward direction of the object). 
Notably, choosing ellipse area ($a>b$) of \textit{yz}-plane is due to that for moving objects on the ground, the motion along the \textit{y}-axis tends to be greater than along the \textit{z} -axis. 
For simplicity, we denote the learned point features $\{{ }^{(T)} \mathbf{F}_{\mathbf{x}_{i}}\}$ and $\{{ }^{(T)} \mathbf{F}_{\mathbf{y}_{j}}\}$ after the last iteration as $\mathbf{F}_{\mathbf{X}}=\{\mathbf{F}_{\mathbf{x}_{i}}\}$ and $\mathbf{F}_{\mathbf{Y}}=\{\mathbf{F}_{\mathbf{y}_{j}}\}$. 

\noindent\textbf{Inlier classifier for outlier rejection.}  
\label{inlier}
Due to the outlier interference (caused by the LiDAR noise or background points in search area), it's difficult to directly predict the precise rigid transformation between the original template and search area. 
To relieve it, we construct an inlier classifier built on a 3-layer MLP for outlier filtering on both the template and search area. 
Specifically, with the learned point feature as input, the classifier predicts the inlier probabilities $\{\hat{c}_{\mathbf{x}_i}\in[0,1]\}$ and $\{\hat{c}_{\mathbf{y}_j}\in[0,1]\}$ for points $\{\mathbf{x}_i\}$ and $\{\mathbf{y}_j\}$. 
Then, we choose $k$ points with the highest inlier probabilities as the inliers, ${\mathbf{X}'}=\{\mathbf{x}_i \mid \hat{c}_{\mathbf{x}_i} \geq \tau_{{x}}\}$ and ${\mathbf{Y}'}=\{\mathbf{y}_i \mid \hat{c}_{\mathbf{y}_i} \geq \tau_y \}$, where $\tau_x$ and $\tau_y$ denote the top-$k$ probability threshold. 
We denote the number of points in the filtered template and search area as $N'=|\mathbf{X}'|$ and $M'=|\mathbf{Y}'|$, respectively. 

After rejecting the outliers, a matching map $\mathbf{M}\in\mathbb{R}^{N'\times M'}$ is established with a softmax function to generate the soft correspondence $\hat{\mathbf{y}}'_i$ for each template point ${\mathbf{x}}'_i$: 
\begin{equation}
	\hat{\mathbf{y}}'_i=\sum_{j=1}^{M'}\mathbf{M}_{i,j}\mathbf{y}'_j,\ \ \ \mathbf{M}_{i,j}=\operatorname{softmax}([\mathbf{F}_{\mathbf{x}'_i}^\top\mathbf{F}_{\mathbf{y}_1'},...,\mathbf{F}_{\mathbf{x}'_i}^\top\mathbf{F}_{\mathbf{y}_{M'}'}])_j. 
\end{equation}
Finally, we use the weighted SVD to solve the least-square fitting over the generated soft correspondence pairs $\{(\mathbf{x}'_i, \hat{\mathbf{y}}'_{i})\}$ for rigid transformation estimation:
\begin{equation}
	\hat{\mathbf{R}},\hat{\mathbf{t}}=\underset{\mathbf{R},\mathbf{t}}{\arg\min} \sum_{i=1} c_{\mathbf{x}'_i}\left\|\mathbf{R}\mathbf{x}'_i + \mathbf{t} -\hat{\mathbf{y}}'_i\right\|, 
\end{equation}
where the inlier probability controls the optimization weight on each least-square item. 
Consequently, we align the template to the search area with the predicted rigid transformation $\{\hat{\mathbf{R}},\hat{\mathbf{t}}\}$, and utilize the transformed template $\bar{\mathbf{X}}=\{ { \bar{\mathbf{x}}_i\mid\hat{\mathbf{R}}}{\mathbf{x}}_i+\hat{\mathbf{t}}\}$ and search area $\mathbf{Y}$ for Siamese object localization. 
We note that the spatially aligned corresponding points via 3D registration tend to achieve consistent feature representation, thereby improving the robustness of the feature matching as shown in Fig~\ref{fig:motivation}. 

\noindent\textbf{Discussion about tracking just using registration.}  
Although our proposed tracking-specific registration module presents high registration precision between the consecutive frames (see Fig.~\ref{reg}), it should be noted that the only-registration based tracker tends to suffer from serious error accumulation issue, resulting in the tracking failure. For example, we assume there is a BBox error $\Delta_t$ between the predicted BBox $\hat{\mathbf{b}}_t$ and the ground-truth BBox ${\mathbf{b}}_t$ in frame $t$ (i.e., ${\mathbf{b}}_t=\hat{\mathbf{b}}_t+\Delta_t$). 
 Then, even if we estimate the perfect relative pose $\mathcal{T}^*$ between frame $t$ and frame $t+1$, the transformed BBox $\hat{\mathbf{b}}_{t+1} = \mathcal{T}^*(\hat{\mathbf{b}}_t)$ still encounters the same BBox error $\Delta_t$ with ground-truth BBox ${\mathbf{b}}_{t+1}$ (i.e., ${\mathbf{b}}_{t+1}={\hat{\mathbf{b}}}_{t+1}+\Delta_t$). 
 Therefore, in the case of imperfect relative pose estimation, the accumulated error tends to increase gradually as the tracking proceeds, causing the poor tracking accuracy. 
Based on the discussion above, we still need the feature aggregation and object localization for 3D tracking as introduced in the next section. 

\subsection{Registration-Aided Sinkhorn Feature Aggregation}
\label{fusion} 
Registration-aided Sinkhorn Feature Aggregation (RSFA) module aims to integrate the target information (defined in the transformed template) into the search area for object localization and mainly contains two components: the Sinkhorn optimization based feature matching and the registration-aided matching refinement. 

\noindent\textbf{Sinkhorn optimization based feature matching.} 
Following previous Siamese methods, we extract the point-wise features $\Phi_{\bar{\mathbf{X}}}=\{\Phi_{\bar{\mathbf{x}}_i}\in\mathbb{R}^d\}$ and $\Phi_{\mathbf{Y}}=\{\Phi_{{\mathbf{y}}_j}\in\mathbb{R}^d\}$  of the transformed template and search area using PointNet++ backbone~\cite{qi2017pointnet++}. 
Different from the previous methods using Cosine distance to measure the feature similarity, we formulate the feature matching as the optimal transport problem, and use the Sinkhorn algorithm~\cite{sinkhorn1964relationship} to search for the reliable similarity assignment, which is verified to be more robust to the outlier interference~\cite{yang2020mapping,yew2020rpm}. 
Specifically, we first construct the initial feature matching matrix $\mathbf{A}\in\mathbb{R}^{N\times M}$ via feature inner product: $\mathbf{A}_{i,j}=<\Phi_{\bar{\mathbf{x}}_i},\Phi_{{\mathbf{y}}_i}>$. 
Since Sinkhorn optimization needs matrix elements to be positive within finite values, we transform the initial matrix via instance normalization and concurrently use an exponential function to map the normalized matrix elements to be positive: $\mathbf{A}^\text{pos}=\operatorname{Exp}(\operatorname{Ins.Norm}(\mathbf{A}))$.
Then, the Sinkhorn algorithm expands the matrix $\mathbf{A}^{\text{pos}}$ with a slack row and a slack column to form a slack matching matrix: 
\begin{equation}
	\mathbf{A}'=\left[\begin{array}{cc}
		\mathbf{A}^{\text{pos}} & \mathbf{z}_1 \\
		\mathbf{z}_2^{T} & z
	\end{array}\right], \quad \mathbf{A}^{\prime} \in \mathbb{R}^{\left(N+1\right) \times\left(M+1\right)},
\end{equation}
where $\mathbf{z}_1\in\mathbb{R}^{M\times1}$, $\mathbf{z}_2\in\mathbb{R}^{N\times1}$ and $z\in\mathbb{R}$ are all optimizable parameters (initialized with zero), providing a similarity assignment space for outliers without corresponding points. 
After repeatedly alternating the row and column normalizations on $\mathbf{A}'$, we remove the slack variables and utilize the resulting matrix $\tilde{\mathbf{A}}\in\mathbb{R}^{N\times M}$ as the feature matching matrix.  

\noindent\textbf{Registration-aided matching refinement.} 
Although the spatial alignment and the Sinkhorn algorithm can effectively improve the matching robustness as above, the feature matching may inevitably suffer from the ambiguous similarity assignment on smooth surfaces that lacks significant geometric characteristic (such as the car-door surface). 
For example, while search-area point $\mathbf{y}_1$ on car-door surface has a high similarity with its aligned corresponding point $\bar{\mathbf{x}}_1$, it may also achieve a certain similarity with another template point $\bar{\mathbf{x}}_2$ also lying on this indistinguishable surface, which potentially degrades the discrimination of $\tilde{\mathbf{A}}$. 
Based on our registration operation, we focus on exploiting the spatial distance constraint of the aligned corresponding points (termed registration matching) to handle this issue. 
Specifically, we first construct a spatial distance-based matching matrix $\mathbf{D}^{{\text{reg}}}\in\mathbb{R}^{N\times M}$ between $\bar{\mathbf{X}}$ and ${\mathbf{Y}}$, and then utilize $\mathbf{D}^{{\text{reg}}}$ to regularize $\tilde{\mathbf{A}}$ using Hadamard product: 
\begin{equation}
	\label{match}
	\begin{split}
		\tilde{\mathbf{A}}_{i,j}^{\text{reg}}&= \tilde{\mathbf{A}}_{i,j}\cdot \mathbf{D}^{\text{reg}}_{i,j}, \\
		\mathbf{D}^{\text{reg}}_{i,j} & = \max\Big(1 -\frac{d^2_{i,j}}{\sigma^2}, 0 \Big),\ \ d_{i,j}=\left\| \bar{\mathbf{x}}_i - \mathbf{y}_j\right\|_2,
	\end{split}
\end{equation} 
where the parameter $\sigma$ controls the maximum allowance spatial distance of the corresponding points. If distance $d_{i,j}$ exceeds the threshold $\sigma$, the entity $\mathbf{D}^{\text{reg}}_{i,j}$ is clipped to 0. Otherwise, the entity $\mathbf{D}^{\text{reg}}_{i,j}$ is negatively related to  $d_{i,j}$. 

Compared to the matching matrix $\tilde{\mathbf{A}}$ just at feature level, the registration-aided spatial distance constraint can effectively relieve the proposed ambiguous matching problem. 
We continue the example presented above to illustrate it. 
After registration, the aligned template point $\bar{\mathbf{x}}_1$ tends to own a much smaller spatial distance with $\mathbf{y}_1$ than $\bar{\mathbf{x}}_2$. Therefore, we can utilize this spatial distance-based matching score to reweight the feature matching score in $\tilde{\mathbf{A}}$ so that the incorrect feature similarity between the non-corresponding pair can be effectively reduced. 

\noindent\textbf{Target-specific feature aggregation.}  
After establishing the feature matching matrix $\tilde{\mathbf{A}}^{\text{reg}}$, we use it to guide the transferring of the target information (defined in the transformed template) into the search area for target-specific search-area feature learning. 
Specifically, we fuse the global target information and the local target information into each search-area point $\mathbf{y}_j$. 
For the global target information, we utilize the matching scores to weight all template features, on which a max-pooling operation is concurrently performed to achieve the global target information. 
A MLP is then used to produce that global target embedding: $\mathbf{F}^g_{\mathbf{y}_j}=\operatorname{MLP}(\operatorname{MaxPool}_i\{\tilde{\mathbf{A}}^{\text{reg}}_{i,j}\cdot\Phi_{\bar{\mathbf{x}}_i}\})$. 
For the local target information, we transfer the local target information of the most relevant template point $\bar{\mathbf{x}}_{k^*}$ (index  $k^*=\arg\max_i \tilde{\mathbf{A}}^{\text{reg}}_{i,j}$) into $\mathbf{y}_j$. The local target information of template-point $\bar{\mathbf{x}}_{k^*}$ consists of the point feature $\Phi_{{\bar{\mathbf{x}}}_{k^*}}$, point coordinate $\bar{\mathbf{x}}_{k^*}$ and the similarity score  $\tilde{\mathbf{A}}^{\text{reg}}_{k,j}$. 
These local target information (combining the search-area feature $\Phi_{{{\mathbf{y}}}_j}$) is then passed into a MLP to build the local target embedding: $\mathbf{F}^l_{\mathbf{y}_j}=\operatorname{MLP}([\Phi_{{{\mathbf{y}}}_j},\tilde{\mathbf{A}}^{\text{reg}}_{k^*,j},\Phi_{\bar{\mathbf{x}}_{k^*}},\bar{\mathbf{x}}_{k^*}])$. 
Finally, with the concated global and local target embeddings, we employ a MLP to fuse them and achieve the desired target-specific feature for point $\mathbf{y}_j$:  $\mathbf{F}^{\text{tgt}}_{\mathbf{y}_j}=\operatorname{MLP}([\mathbf{F}^g_{\mathbf{y}_j},\mathbf{F}^l_{\mathbf{y}_j}])$. 
Finally, we feed the target-specific features $\left\{\mathbf{F}^{\text{tgt}}_{\mathbf{y}_j}\right\}$ in a CenterPoint-like detection head~\cite{yin2021center,hui20213d} for object center and rotation-angle regression. 

\subsection{Loss Function}
The loss functions of our method consist of the registration-level loss and the tracking-level loss. 
Given the template $\mathbf{X}$ with BBox $\left(x_1,y_1,z_1,\theta_1\right)$ and search area $\mathbf{Y}$ with BBox $\left(x_2,y_2,z_2,\theta_2\right)$, we construct the ground-truth rotation matrix $\mathbf{R}^*$ and translation vector $\mathbf{t}^*$ for registration supervision, 
\begin{equation}
	\begin{split}
		{\mathbf{R}^*}=\left[\begin{array}{ccc}
			\operatorname{cos}(\Delta\theta) & -\operatorname{sin}(\Delta\theta) & 0 \\
			\operatorname{sin}(\Delta\theta)  & \operatorname{cos}(\Delta\theta) & 0 \\
			0 & 0 & 1
		\end{array}\right], \ \ 
		{\mathbf{t}^*} = \left[\Delta x, \Delta y, \Delta z\right], 
	\end{split}
\end{equation}
where $\Delta\theta = \theta_2 - \theta_1$ and $\Delta x = x_2 - x_1$. 
The ground-truth transformed template is denoted as $\tilde{\mathbf{X}}=\{ \tilde{\mathbf{x}}_i \in \mathbb{R}^3 \mid {\tilde{\mathbf{x}}_i=\mathbf{R}^*}{\mathbf{x}}_i+\mathbf{t}^*, \mathbf{x}_i\in\mathbf{X}\}$. 

\noindent\textbf{Inlier classification supervision.} 
To train our inlier classifier, we need prepare the inlier/outlier labels for all template and search-area points (denoted as $\{c^*_{\mathbf{x}_i}\}$ and $\{c^*_{\mathbf{y}_j}\}$) which are generated via: 
\begin{equation}
	\begin{split}
		c^*_{\mathbf{x}_i} = \mathbb{I}\{d_{\tilde{\mathbf{x}}_i\rightarrow\mathbf{Y}}<\tau \},\ \ 
		c^*_{\mathbf{y}_j} = \mathbb{I}\{d_{{\mathbf{y}_j}\rightarrow\tilde{\mathbf{X}}}<\tau \},
	\end{split}
\end{equation}
where  $d_{\tilde{\mathbf{x}}_i\rightarrow\mathbf{Y}}=\min_{\mathbf{y}_j}\left\|\tilde{\mathbf{x}}_i- \mathbf{y}_j\right\|_2$ denotes the 
minimum $l_2$ distance from $\tilde{\mathbf{x}}_i$ to $\mathbf{Y}$ and $\tau$ is the inlier threshold (we set $\tau$ to 0.1 for all our experiments). We adopt the binary cross-entropy (BCE) loss as the supervision signal: 
\begin{equation}
	\begin{split}
		\mathcal{L}_{\text{cls}} = \sum_{i=1}\operatorname{BCE}\left(\hat{c}_{\mathbf{x}_i}, c^*_{\mathbf{x}_i}\right) + \sum_{j=1}\operatorname{BCE}\left(\hat{c}_{\mathbf{y}_j}, c^*_{\mathbf{y}_j}\right),
	\end{split}
\end{equation}
where $\operatorname{BCE}(p,q)=-q\log(p)- (1-q)\log(1-p)$. 

\begin{table*}[h]
	\centering
	\resizebox{1.0\linewidth}{!}{
		\begin{tabular}{c|c|ccccc|ccccc}
			\toprule[1.5pt]
			& Metrics & \multicolumn{5}{c|}{\emph{Success}}& \multicolumn{5}{c}{\emph{Precision}}\\
			\hline
			& Category & Car & Pedestrian & Van & Cyclist & Mean  & Car & Pedestrian & Van & Cyclist & Mean \\
			& Frame Num. & 6424 & 6088 & 1248 & 308 & 14068 & 6424 & 6088 & 1248 & 308 & 14068 \\
			\hline
			\multirow{8}*{\rotatebox{90}{KITTI}} & SC3D~\cite{giancola2019leveraging} & 41.3 & 18.2 & 40.4 & 41.5 & 31.2 & 57.9 & 37.8 & 47.0 & 70.4 & 48.5\\
			& P2B~\cite{qi2020p2b} & 56.2 & 28.7 & 40.8 & 32.1 & 42.4 & 72.8 & 49.6 & 48.4 & 44.7 & 60.0\\
			& LTTR~\cite{cui3d} & 65.0 & 33.2 & 35.8 & \underline{66.2} & 48.7 & 77.1 & 56.8 & 45.6 & 89.9 & 65.8\\
			& BAT~\cite{zheng2021box} & 60.5 & 42.1 & 52.4 & 33.7 & 51.2 & 77.7 & 70.1 & \underline{67.0} & 45.4 & 72.8\\
			& PTT~\cite{shan2021ptt} & 67.8 & 44.9 & 43.6 & 37.2 & 55.1 & \underline{81.8} & 72.0 & 52.5 & 47.3 & 74.2 \\
			& PTTR~\cite{zhou2021pttr} & 65.2 & \underline{50.9} & \underline{52.5} & 65.1 & \underline{58.4}  & 77.4 & \underline{81.6} & 61.8 & \underline{90.5} & \underline{77.8}\\
			& V2B~\cite{hui20213d} & \underline{70.5} & 48.3 & 50.1 & 40.8 & \underline{58.4}  & 81.3 & 73.5 & 58.0 & 49.7 & 75.2\\
			\cmidrule{2-12}
			& RDT (ours) & {\bf 71.8}  & {\bf 56.4}  & {\bf 60.4} & {\bf 72.8} & {\bf 64.1} & {\bf 83.2} & {\bf 84.1} & {\bf 69.7} & {\bf 93.7} & {\bf 82.6}\\ 
			\midrule[1pt]
			& Category & Car & Pedestrian & Truck & Bicycle & Mean  & Car & Pedestrian & Truck & Bicycle & Mean \\
			& Frame Num. & 15578 & 8019 & 3710 & 501 & 27808 & 15578 & 8019 & 3710 & 501 & 27808 \\
			\hline
			\multirow{5}*{\rotatebox{90}{NuScenes}} & SC3D~\cite{giancola2019leveraging} & 25.0 & 14.2 & 25.7 & 17.0 & 21.8 & 27.1 & 16.2 & 21.9 & {18.2} & 23.1 \\
			&P2B~\cite{qi2020p2b} & 31.8 & 18.7 & 21.7 & 17.8 & 26.4 & 33.6 & 24.7 & 17.8 & {21.9} & 28.7\\
			& BAT~\cite{zheng2021box} & 30.1 & {18.9} & 23.5 & 16.5 & 25.7 & 31.3 & {26.0} & 18.5 & 19.2 & 27.8\\
			& V2B~\cite{hui20213d} & \underline{36.6} & \underline{19.3} & \underline{31.5} & \textbf{18.9} & \underline{30.6} & \underline{39.2} & \underline{26.6} & \underline{29.3} & \underline{22.0} & \underline{33.9}\\
			\cmidrule{2-12}
			& RDT (ours) & \textbf{37.1} & \textbf{20.4} & \textbf{33.4} & \underline{18.1} & \textbf{31.4} & \textbf{39.8} & \textbf{29.0} & \textbf{30.2} & \textbf{22.9}& \textbf{35.1}\\ 			
			\bottomrule[1.5pt]
		\end{tabular}
	}
	\vspace{1mm}
	\caption{Performance comparisons with SOTA trackers on \textit{Car}, \textit{Pedestrian}, \textit{Van} and \textit{Cyclist} categories from \textit{KITTI} and \textit{NuScenes} datasets.
	}
	\label{tab:base_results}
\end{table*}

\noindent\textbf{Correspondence supervision.} 
The correspondence supervision aims to improve the quality of the generated soft correspondence $\left\{\hat{\mathbf{y}}'_i\right\}$ for inlier-filtered template points $\left\{{\mathbf{x}}'_i\right\}$ (see Sec.~\ref{inlier}) in our registration module. 
We realize it by minimizing the $l_2$ loss between the soft correspondence and the ground-truth transformed template point as below:
\begin{equation}
	\begin{split}
		\mathcal{L}_{\text{corr}} = \sum_{i=1}\left\| \mathbf{R}^*{\mathbf{x}}'_i+\mathbf{t}^* - \hat{\mathbf{y}}'_i\right\|_2.
	\end{split}
\end{equation}

\noindent\textbf{Registration supervision.} 
We utilize the transformation loss $\mathcal{L}_{\operatorname{trans}}$ as below to promote the predicted transformation to approach the ground-truth transformation,
\begin{equation}
	\begin{split}
		\mathcal{L}_{\text{trans}} = \left\|\hat{\mathbf{R}}^\top\mathbf{R}^*-\mathbf{I}\right\|^2 + \left\|\hat{\mathbf{t}}-\mathbf{t}^*\right\|_2^2.
	\end{split}
\end{equation}
The final loss function combines the registration-level losses above and the tracking-level losses $\mathcal{L}_{\text{track}}$ (including the center-coordinate and rotation-angle regression losses as in \cite{hui20213d,yin2021center}), 
\begin{equation}
	\begin{split}
		\mathcal{L} = \mathcal{L}_{\text{cls}} + \mathcal{L}_{\text{corr}} + \mathcal{L}_{\text{trans}} + \mathcal{L}_{\text{track}}. 
	\end{split}
\end{equation}

\begin{figure*}[t]
	\centering
	\includegraphics[width=1.0\textwidth]{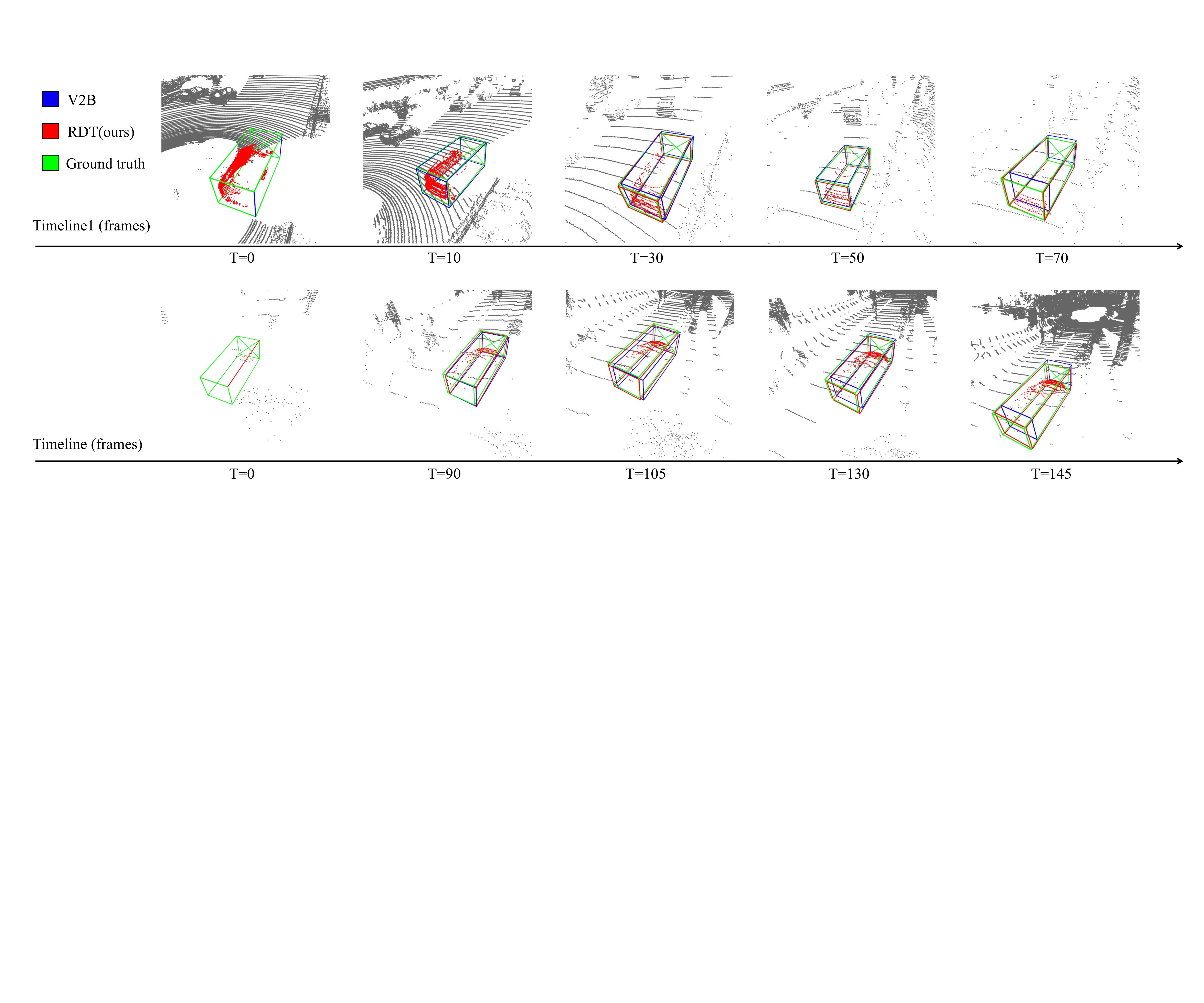}
	\caption{Qualitative comparison with the SOTA tracker V2B on \textit{Car} category from \textit{KITTI} dataset.}
	\label{vis}
\end{figure*}

\begin{figure*}[t]
	\centering
	\includegraphics[width=1.0\textwidth]{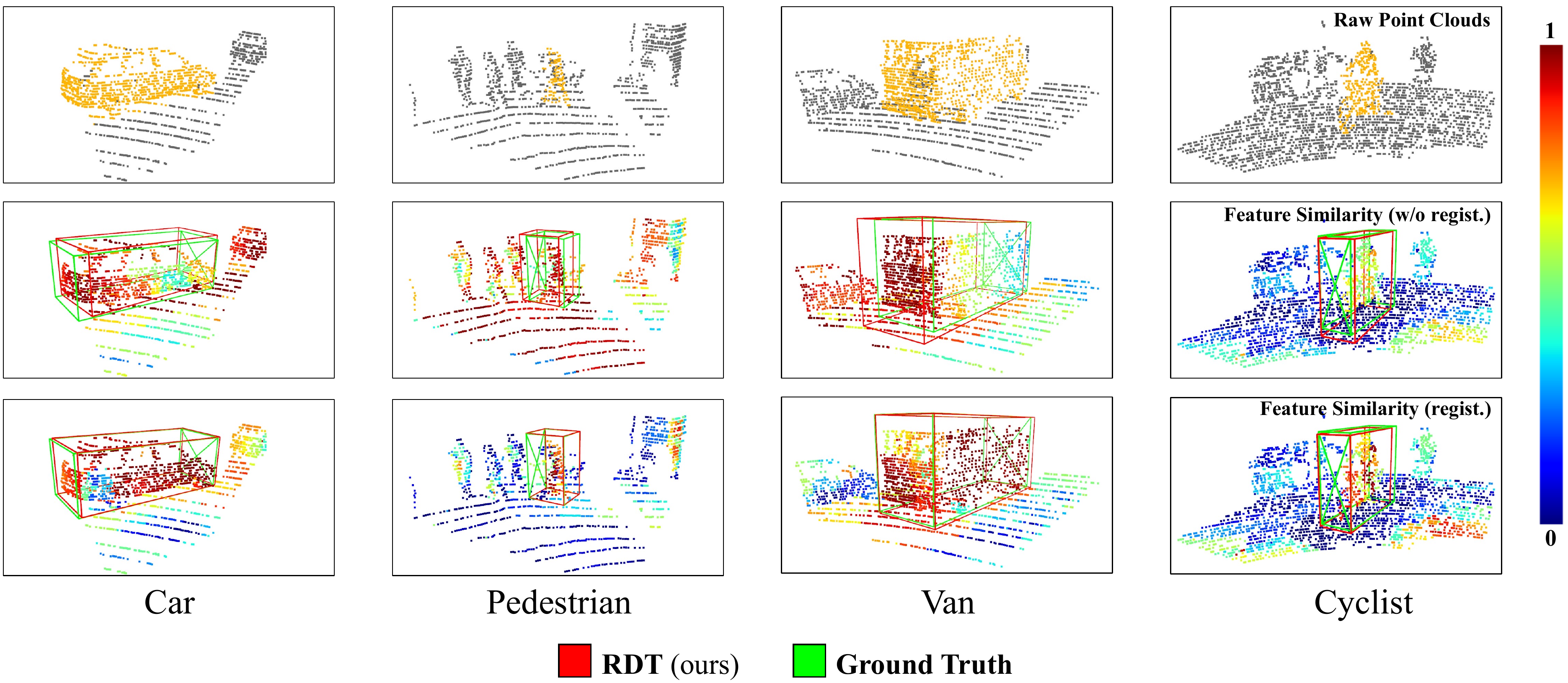}
	\caption{Feature similarity visualization on \textit{KITTI} dataset. The top row is the raw point cloud; The second row is the feature similarity without the registration module; The third row is the feature similarity with the registration module. After performing the registration between the template and the search area, their corresponding points can achieve much higher feature similarity.  
	}
	\label{vis2}
\end{figure*}

\section{Experiments}
To verify the effectiveness of our method, we perform extensive experiments on different benchmark datasets and ablation studies in this section. 
In concrete, we first present our implementation details, the training and testing processing, and the evaluation metric in Sec.~\ref{setting}. 
Then, we compare our method to some state-of-the-art (SOTA) methods on \textit{KITTI}, \textit{NuScenes} and  \textit{Waymo} datasets in Sec.~\ref{eval_kitti}, Sec.~\ref{eval_nuscenes} and Sec.~\ref{eval_waymo}, respectively. Finally, we present our extensive ablation studies about the proposed modules and the hyper-parameter settings in Sec.~\ref{ablation}. 

\begin{table*}
	\centering
	\resizebox{0.8\textwidth}{!}{
		\begin{tabular}{c|c|cccc|cccc}
			\toprule[1.5pt]
			& Category & \multicolumn{4}{c|}{Vehicle}& \multicolumn{4}{c}{Pedestrian}\\
			\midrule
			& Split & Easy & Medium & Hard & Mean & Easy & Medium & Hard & Mean \\
			& Frame Num. & 67832 & 61252 & 56647 & 185731 & 85280 & 82253 & 74219 & 241752 \\
			\midrule
			\multirow{4}*{\rotatebox{90}{\emph{Success}}} & P2B~\cite{qi2020p2b} & 57.1 & 52.0 & 47.9 & 52.6 & 18.1 & 17.8 & 17.7 & 17.9\\
			& BAT~\cite{zheng2021box} & 61.0 & 53.3 & 48.9 & 54.7 & 19.3 & 17.8 & 17.2 & 18.2\\
			& V2B~\cite{hui20213d} & \underline{64.5} & \underline{55.1} & \underline{52.0} & \underline{57.6} & \underline{27.9} & \underline{22.5} & \underline{20.1} & \underline{23.7}\\
			\cmidrule{2-10}
			& RDT (ours) & \textbf{65.5} & \textbf{58.8} & \textbf{52.4} & \textbf{58.3} & \textbf{28.6} & \textbf{24.2} & \textbf{21.3} & \textbf{24.9}\\ 
			\midrule
			\multirow{4}*{\rotatebox{90}{\emph{Precision}}} & P2B~\cite{qi2020p2b} & 65.4 & 60.7 & 58.5 & 61.7 & 30.8 & 30.0 & 29.3 & 30.1\\
			& BAT~\cite{zheng2021box} & 68.3 & 60.9 & 57.8 & 62.7 & 32.6 & 29.8 & 28.3 & 30.3\\
			& V2B~\cite{hui20213d} & \underline{71.5} & \underline{63.2} & \underline{62.0} & \underline{65.9} & \underline{43.9} & \underline{36.2} & \underline{33.1} & \underline{37.9}\\
			\cmidrule{2-10}
			& RDT (ours) & \textbf{72.3} & \textbf{64.0} & \textbf{62.8} & \textbf{66.5} & \textbf{45.2} & \textbf{38.7} & \textbf{35.2} & \textbf{39.9}\\ 
			\bottomrule[1.5pt]
		\end{tabular}
	}
	\vspace{1.5mm}
	\caption{The performance of different methods on the waymo open dataset. Each category is divided into three levels of difficulty: Easy, Medium and Hard. ``Mean'' denotes the average results of three difficulty.}
	\label{tab:waymo_results}
\end{table*}

\subsection{Experimental Settings}
\label{setting}
\noindent\textbf{Implementation details.}  
In our registration module, we set nonlocal iteration times $T$ and feature dimension $d$ to 12 and 128. 
The semi-major and semi-minor axes $a$ and $b$ in Eq.~\ref{beta} are set to 1.6 and 0.4. 
We choose $50\%$ points with the highest confidence for inlier selection. 
In the feature aggregation module, the maximum allowance spatial distance $\sigma$ is set to 0.4. 
We train the model in 30 epochs, where the first 5 epochs are used to train the registration network only, and the remaining 25 epochs are used to jointly train the registration and tracking modules.
We use the optimizer Adam with a learning rate 0.001 and weight decay by 0.2 every 6 epochs for model training. 
We implement our model with \textit{PyTorch} and deploy all experiments on a server containing an Intel i5 2.2 GHz CPU and two TITAN RTX GPUs with almost 24 GB per card. 
For simplicity, we name our \textbf{R}egistration-\textbf{D}riven \textbf{T}racker as \textbf{RDT}. 

\noindent\textbf{Training and testing.}  
For training, we augment the training samples by adding random offsets on the previous and current ground-truth BBoxes.
The former is used to crop the template, and the latter is enlarged by 2 meters for search area cropping.
For testing, we fuse the points within the first and previous BBoxes to generate the template, while the search area is cropped by the previous BBox enlarged by 2 meters.
The template and search area points are set to 512 and 1024. 

\noindent\textbf{Evaluation Metrics.}  
Following~\cite{qi2020p2b},  we use \textit{Success} and \textit{Precision} criteria by one-pass evaluation (OPE) to measure the model performance. 
\textit{Success} is defined as the intersection over
union (IoU) between the predicted and the ground-truth BBoxes and \textit{Precision} is the AUC score (area under curve) for the distance between their centers from 0 to 2m.

\subsection{Evaluation on KITTI dataset}
\label{eval_kitti}
We first evaluate our method on \textit{KITTI} benchmark~\cite{geiger2012we}, a LiDAR-scanned driving scenarios dataset, including 21 outdoor scenes and 8 object categories. 
For a fair comparison, following the processing and data split in~\cite{qi2020p2b}, we utilize the sequences 0-16 scenes, 17-18 scenes, and 19-20 scenes for training, validation, and testing, respectively. 
We compare our method to seven SOTA trackers, i.e., SC3D~\cite{giancola2019leveraging}, P2B~\cite{qi2020p2b}, 3DSiamRPN~\cite{fang20203d}, BAT~\cite{zheng2021box}, PTT~\cite{shan2021ptt}, PTTR~\cite{zhou2021pttr} and V2B~\cite{hui20213d}. 
The performance of our method is shown in Table ~\ref{tab:base_results}. 
Overall, our method can achieve an impressive performance advantage on both \textit{Success} and \textit{Precision} criteria compared to other trackers. 
Notably, compared to V2B that also uses the CenterPoint-like detection head like us, our method presents an superior performance on all categories, such as \textit{Pedestrian} category (\textit{Succ}: 48.3$\rightarrow$56.4, \textit{Prec}: 73.5$\rightarrow$84.1). 
This performance gain mainly benefits from our robust feature matching between the template and search area after registration operation. 
Also, the proposed registration-aided Sinkhorn feature aggregation can efficiently transfer the target information into the search area to generate the discriminative target-specific feature for reliable object localization. 

\begin{figure*}[t]
	\centering
	\includegraphics[width=0.9\textwidth]{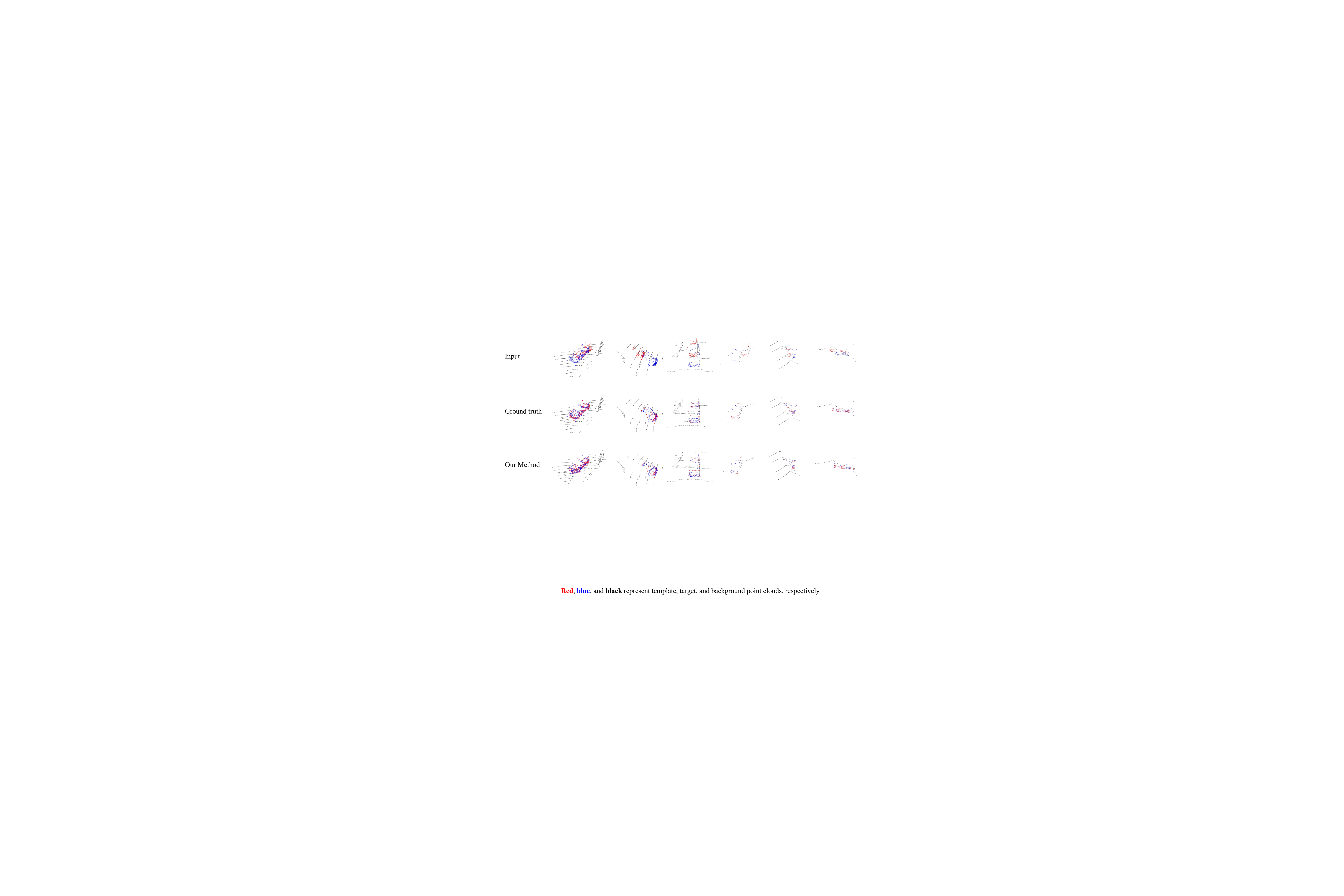}
	\caption{Visualization results about the rigid registration between the template (\textcolor{red}{red}) and search area (\textcolor{blue}{blue}+\textcolor{gray}{grey})  on \textit{Car} category from \textit{KITTI} dataset.}
	\label{reg}
\end{figure*}

\vspace{-2mm}
\subsection{Evaluation on NuScenes dataset}
\label{eval_nuscenes} 
We further test our method on the \textit{NuScenes} benchmark~\cite{caesar2020nuscenes}, which is also an outdoor dataset with 1000 driving scenes and 23 annotated object categories.
Following~\cite{hui20213d}, we split it into 750 training sequences and 150 validation sequences, where we only evaluate the trained model on the latter since the annotations for its test split are not accessible. 
We compare our tracker to three representative SOTA tackers, i.e., P2B, BAT, and V2B.  
As shown in Table ~\ref{tab:base_results}, 
our method can still outperform all trackers on both \textit{Success} and \textit{Precision} criteria in most categories. 
Notably, different from the 64-beam LiDAR used in \textit{KITTI} dataset, \textit{NuScenes} is just scanned by the 32-beam. 
Therefore, the existing data gap between the \textit{KITTI} and the NuScenes largely degrades our generalization precision and RDT just can achieve limited performance improvement. 

\subsection{Evaluation on Waymo  dataset}
\label{eval_waymo} 
\textit{Waymo} dataset~\cite{sun2020scalability} is a large-scale outdoor dataset containing 150 scenes for tesing, where the amount of frames and the scene complexities are significantly beyond the \textit{KITTI} and \textit{NuScenes} datasets.  
To evaluate the generalization ability of our method, we directly generalize the model learned by \textit{KITTI} to the \textit{Waymo} dataset. As shown in Table~\ref{tab:waymo_results}, we test on Vehicle and Pedestrian categories and our method outperforms other methods (P2B, BAT, and V2B) in terms of different difficulty levels (easy, medium, and hard), which demonstrates the strong generalization ability of our model. 

\begin{table}[]
	\centering
	\resizebox{0.95\columnwidth}{!}{
		\begin{tabular}{ccccc|cc}
			\toprule[1.5pt]
			\textit{TSNR} & \textit{TSNonlocal} & \textit{Classifier}  & \textit{RM} & \textit{Sinkhorn} &Succ.& Prec.  \\ \midrule 
			& & &  & $\surd$  &  {68.7} & {78.7} \\ 
			$\surd$ & &$\surd$ & $\surd$ & $\surd$  & \underline{70.3} & {80.9} \\ 
			$\surd$&$\surd$&&$\surd$& $\surd$& 69.7 & 80.2  \\
			$\surd$& $\surd$ & $\surd$ & & $\surd$ &  {69.4} & {79.7} \\ 
			$\surd$& $\surd$ & $\surd$ & $\surd$  & &  {70.1} & \underline{82.0} \\ 
			$\surd$ &$\surd$  &$\surd$ &$\surd$ & $\surd$ & \textbf{71.8} & \textbf{83.2} \\ 
			\bottomrule[1.5pt]
	\end{tabular}}
	\vspace{1mm}
	\caption{Ablation studies of different components on \textit{Car} catergory from \textit{KITTI} dataset. \textit{TSNR}: Tracking-specific nonlocal registration; \textit{TSNonlocal}: Spatial distance constrained nonlocal module; \textit{Classifier}: Inlier classifier; \textit{RM}: Registration matching; \textit{Sinkhorn}: Sinkhorn optimization.}
	\label{different2}
\end{table}

\subsection{Ablation Study} 
\label{ablation}
\noindent\textbf{Registration module and its components.}  
To verify the effectiveness of our Tracking-Specific Nonlocal Registration (\textit{TSNR}) module, we remove it (its related components containing \textit{TSNonlocal}, \textit{Classifier} and \textit{RM} are also removed) and test its performance on the representative \textit{Car} category from \textit{KITTI} dataset. 
As shown in the first row of Table~\ref{different2}, without \textit{TSNR}, the \textit{Success} and \textit{Precision} scores of our method degrade significantly (\textit{Succ}: 71.8$\rightarrow$68.7 and \textit{Prec}: 83.2$\rightarrow$78.7). 
Furthermore, we test the performance gain from \textit{TSNonlocal} by replacing it with a traditional nonlocal module without spatial distance constraint $\beta_{i, j}$ in Eq.~\ref{nonlocal_eq}. 
The second row of Table~\ref{different2} shows that the used spatial distance constraint can bring 1.5 and 2.3 performance gain on \textit{Success} and \textit{Precision} criteria, respectively, which demonstrates that enhancing our nonlocal module by tracking-specific cross-attention regularization is effective. 
Moreover, we test the inlier classifier (\textit{Classifier}) in \textit{TSNR}. 
The third row of Table~\ref{different2} shows that outlier rejection can bring impressive gain for our method since the outliers may largely interfere the registration and thus degrade the final tracking performance.

\noindent\textbf{Registration-aid Sinkhorn feature aggregation.}  
The Registration-aided Sinkhorn Feature Aggregation (\textit{RSFA}) module focuses on using the Sinkhorn optimization (\textit{Sinkhorn}) and the registration-aided matching refinement (i.e., Registration Matching (\textit{RM})) to correct the feature matching matrix for robust target-specific feature aggregation. 
The fourth row of Table~\ref{different2} shows that \textit{RM} can bring almost \textit{Succ}: 2.4 and \textit{Prec}: 3.5 performance gain, which verifies that registration-aided spatial distance map facilitates a more reliable feature matching construction for target-specific feature learning. 
In addition, as shown in the fifth row of Table~\ref{different2}, by replacing the traditional Cosine distance-based feature matching, the Sinkhorn optimization-based matching map can also achieve a significant performance improvement, which is mainly due to the robustness to outliers of the Sinkhorn feature matching. 

\begin{figure}[t]
	\centering  
	\subfigure[Semi-major axis $a$.]{
		\label{summary_3}
		\includegraphics[width=0.48\columnwidth]{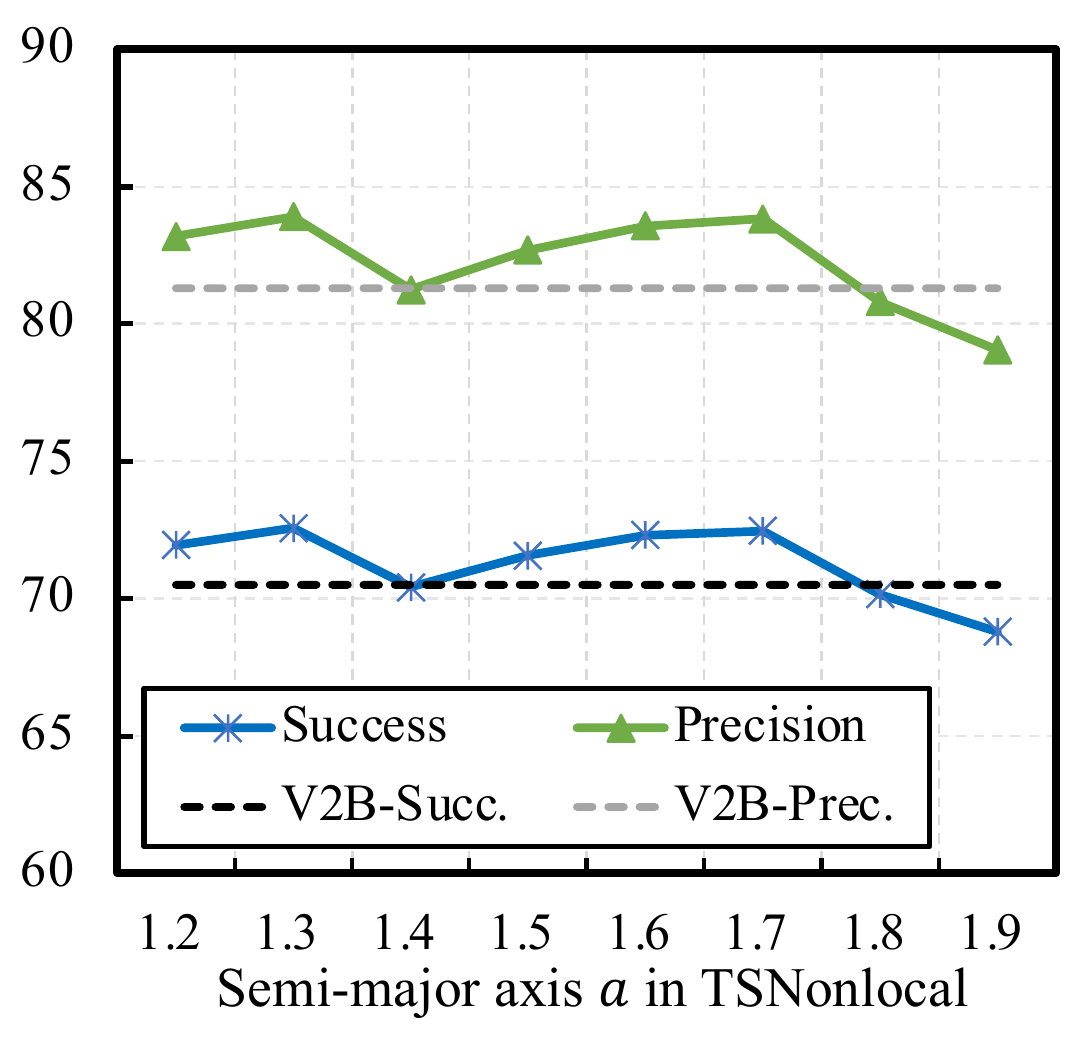}}
	\subfigure[ Maximum allowance $\sigma$.]{
		\label{summary_4}
		\includegraphics[width=0.47\columnwidth]{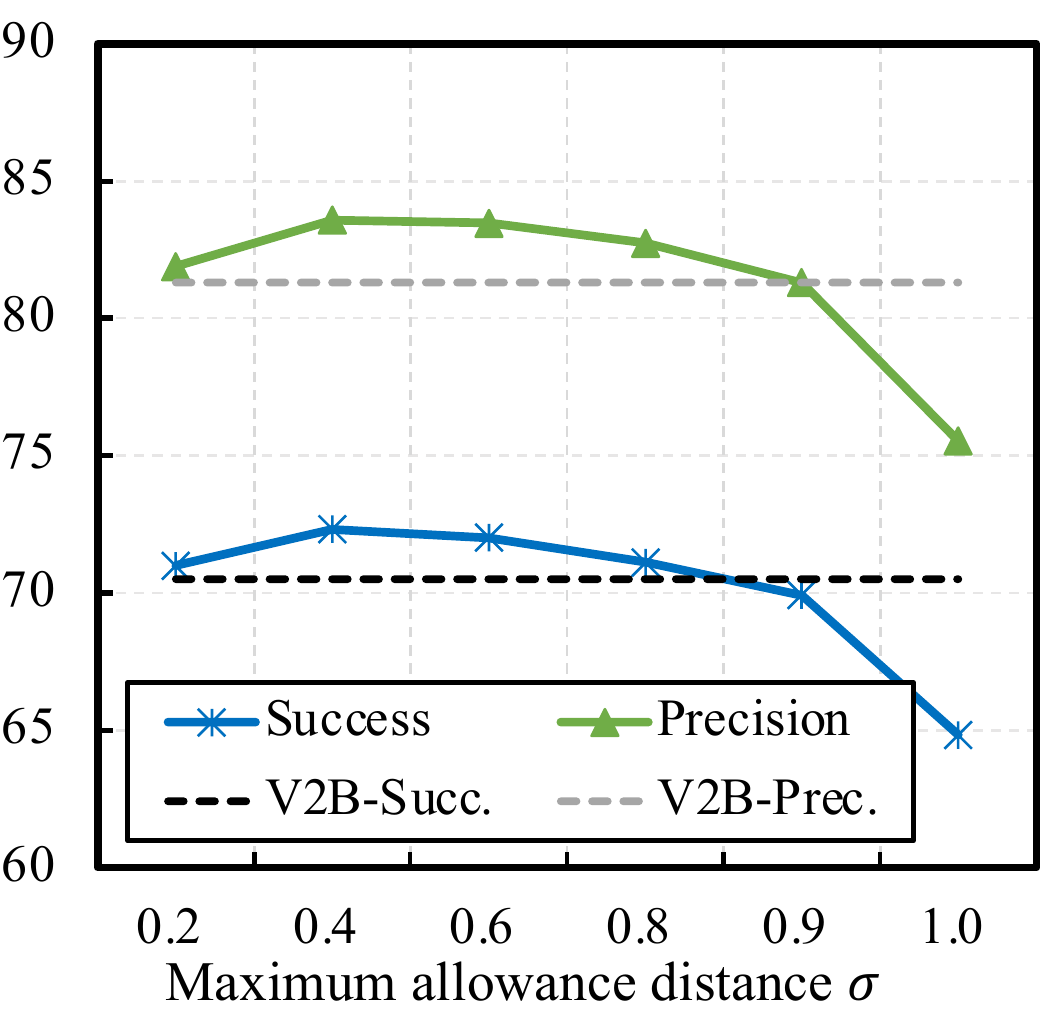}}
	\caption{\textit{Success} and \textit{Precision} in the cases of (a) different semi-major axis $a$ in \textit{TSNonlocal} module, and (b) different maximum allowance distances $\sigma$ in \textit{RM} module on \textit{Car} from \textit{KITTI} benchmark dataset.}
	\label{parameter}
\end{figure}

\noindent\textbf{Parameter setting.}  
We further evaluate the performance changes under different settings of semi-major axis $a$ of \textit{TSNonlocal} (Eq.~\ref{beta}) and maximum allowance distance $\sigma$ of \textit{RM} in Fig.~\ref{parameter} (Eq.~\ref{match}). 
It's noted that we directly generalize the pre-trained model to these parameters without re-training. Compared to the SOTA V2B, our method outperforms it in most settings without specific training, which demonstrates the robustness of our model on parameter selection. 

\begin{table*}[htbp]
	\centering
	\vspace{50pt}
	\resizebox{0.75\textwidth}{!}{
		\begin{tabular}{c|ccccc}
			\toprule[1.5pt]
			Method & Car & Pedestrian & Van & Cyclist & Mean \\
			Total Frame Number & 6424 & 6088 & 1248 & 308 & 14068 \\
			\midrule
			Point Interval & [0, 150) & [0, 100) & [0, 150) & [0, 100) & \\
			Frame Number & 3293 & 1654 & 734 & 59 &  5740\\
			\midrule
			SC3D~\cite{giancola2019leveraging}& 37.9 / 53.0 & 20.1 / 42.0 & 36.2 / 48.7 & \underline{50.2} / \underline{69.2} & 32.7 / 49.4 \\
			P2B~\cite{qi2020p2b}& 56.0 / 70.6 & 33.1 / 58.2 & 41.1 / 46.3 & 24.1 / 28.3 & 47.2 / 63.5 \\
			BAT~\cite{zheng2021box}& 60.7 / 75.5 & 48.3 / \underline{77.1} & 41.5 / 47.4 & 25.3 / 30.5 & 54.3 / 71.9 \\
			V2B~\cite{hui20213d} & \textbf{64.7} / \textbf{77.4} & \underline{50.8} / 74.2 & \underline{46.8} / \underline{55.1} & 30.4 / 37.2 & \underline{58.0} / \underline{73.2} \\
			\cmidrule{1-6}
			RDT (ours) & \underline{64.0} / \underline{77.3} & \textbf{57.2} / \textbf{86.4} & \textbf{55.0} / \textbf{65.4} & \textbf{70.0} / \textbf{91.5} & \textbf{60.9} / \textbf{78.5}\\
			\midrule
			Point Interval & [150, 1000) & [100, 500) & [150, 1000) & [100, 500) & \\
			Frame Number & 2156 & 3112 & 333 & 145 &  5746\\
			\midrule
			SC3D~\cite{giancola2019leveraging}& 36.1 / 53.1 & 17.7 / 38.2 & 38.1 / 53.3 & \underline{44.7} / \underline{76.0} & 26.5 / 45.6 \\
			P2B~\cite{qi2020p2b}& 62.3 / 78.6 & 25.1 / 46.0 & 41.7 / 50.5 & 35.4 / 46.5 & 40.3 / 58.5 \\
			BAT~\cite{zheng2021box} & 71.8 / 83.9 & 45.0 / 71.2 & 44.0 / 51.6 & 41.5 / 52.2 & 54.8 / 74.3 \\
			V2B~\cite{hui20213d} & \underline{77.5} / \underline{87.1} & \underline{46.8} / \underline{72.0} & \underline{51.2} / \underline{59.6} & 44.4 / 53.9 & \underline{58.5} / \underline{76.5} \\
			\cmidrule{1-6}
			RDT (ours) & \textbf{78.4} / \textbf{87.2} & \textbf{56.2} / \textbf{84.4} & \textbf{64.8} / \textbf{73.7} & \textbf{75.5} / \textbf{94.6} & \textbf{65.5} / \textbf{85.0}\\
			\bottomrule[1.5pt]
		\end{tabular}
	}
	\vspace{1.5mm}
	\caption{The results of \emph{Success/Precision} at different point intervals in \textit{KITTI} benchmark dataset. 
	}
	\label{tab:diff_intervals}
\end{table*}


\noindent\textbf{Feature similarity visualization.} In Fig.~\ref{vis2}, we visualize the feature similarity scores of the corresponding points between the template and the search area with/without registration module in four categories of \textit{KITTI} dataset. 
It can be observed that the feature similarities without the registration module are ambiguous. For example, in \textit{Car} category, the template achieves high feature similarities with the background points (gray points in the top row) due to the similar geometric structure with the background car. 
Instead, the registration-enhanced similarity map can effectively relieve it and provide more robust feature matching for reliable object localization. 

\noindent\textbf{Qualitative comparison and registration visualization.} 
We show some representative visual comparisons with V2B on the \textit{Car} category of \textit{KITTI} in Figure ~\ref{vis}, where the top and bottom sequences show the dense-to-sparse and sparse-to-dense sequences, respectively. 
It can be observed that our method can handle well in the challenging scenes with sparse LiDAR scanning. 
More visualization results can be seen in our Appendix II. 
In addition, we also show the registration results between the template and the search area in Fig.~\ref{reg}. 
Due to our tracking-specific registration design, our method can precisely align the template to search area, whether in dense scenes (columns 1-3) or sparse scenes (columns 4-6). 
We also present more registration visualization results in Appendix III. 

\noindent\textbf{Performance evaluation in sparse scenes.} 
To further evaluate our tracking robustness in sparse scenes, we report the performance comparison at difference point intervals of \textit{KITTI} dataset in Table~\ref{tab:diff_intervals}. It can be observed that our model can still significantly outperform other SOTA trackers under different sparse levels (except for point interval [0, 150) in catergory $Car$). 
Particularly, compared to V2B, our model can even achieve $7\%$ and $8.5\%$ performance gain on the averaged \textit{Success} and \textit{Precision} scores, respectively. 
The experimental results on more point intervals can be seen in Appendix I.

\noindent\textbf{Inference speed.}  
We use FPS for inference speed evaluation. 
For a fair comparison, we run each tracker on \textit{Car} class from \textit{KITTI} with the same configuration (a server with a TITAN RTX GPU).
The tracking speed of our method is 13 FPS, and SC3D, P2B, BAT, 3DSiamRPN, PTT, PTTR and V2B can achieve 3 FPS, 28 FPS, 30 FPS, 12 FPS, 24 FPS, 31 FPS and 22 FPS, respectively. 
Although our tracking speed tends to be slower than P2B, BAT, PTT, PTTR and V2B due to the time cost of the registration, the registration module can bring a significant performance gain at a tolerable speed. 

\section{Conclusion}
In this paper, we proposed a novel point cloud registration-driven robust feature matching framework for 3D single object Siamese tracking. 
Our framework includes two key components, including the tracking-specific nonlocal registration module and the registration-aided Sinkhorn feature aggregation module. 
The former aims to construct the reliable feature matching by spatially aligning the template to the search area, where the spatial distance constraint of the corresponding points is integrated into its nonlocal module for robust feature learning. 
The latter combines the Sinkhorn optimization and the registration-aided spatial distance map to establish the robust feature matching map. 
Finally, the constructed feature matching map guides the transferring of the target information into the search area for object localization.
Extensive experiments on benchmark datasets demonstrate the effectiveness of our proposed method.

	\bibliographystyle{IEEEtran}
	\bibliography{IEEEfull}

\end{document}